\documentclass[letterpaper]{article} 
\usepackage[preprint]{aaai2027}  
\usepackage[hyphens]{url}  
\usepackage{graphicx} 
\usepackage{natbib}  
\usepackage{caption} 
\usepackage{algorithm}
\usepackage{algorithmic}

\usepackage{xcolor}

\definecolor{DarkRed}{RGB}{160,0,0}
\definecolor{DarkBlue}{RGB}{0,55,140}

\newcommand{\best}[1]{{\color{DarkRed}\bfseries #1}}
\newcommand{\second}[1]{{\color{DarkBlue}#1}}
\newcommand{\third}[1]{\underline{#1}}

\usepackage{newfloat}
\usepackage{listings}
\DeclareCaptionStyle{ruled}{labelfont=normalfont,labelsep=colon,strut=off} 
\floatstyle{ruled}
\newfloat{listing}{tb}{lst}{}
\floatname{listing}{Listing}

\usepackage{amsmath,amssymb}
\usepackage{booktabs}
\usepackage{array}

\usepackage{multirow}
\newcommand{\sidegroup}[2]{%
  \multirow{#1}{=}{\centering\arraybackslash #2}%
}

\title{ViTexSZ: Heterogeneous Vision--Text Knowledge Distillation for EEG\\Seizure Detection}

\author{
    Chenxi Liu\textsuperscript{\rm 1},
    Mingzhao Li\textsuperscript{\rm 1},
    Yicong Liu\textsuperscript{\rm 2},
    Hao Miao\textsuperscript{\rm 3},\\
    Hongyuan Zhang\textsuperscript{\rm 1},
    Ziyi Chen\textsuperscript{\rm 2},
    Gaofeng Meng\textsuperscript{\rm 1,\rm 4,\rm 5}
}
\affiliations{
    \textsuperscript{\rm 1}Centre for Artificial Intelligence and Robotics, Hong Kong Institute of Science \& Innovation,\\Chinese Academy of Sciences, Hong Kong SAR\\
    \textsuperscript{\rm 2}The First Affiliated Hospital, Sun Yat-sen University, China\\
    \textsuperscript{\rm 3}The Hong Kong Polytechnic University, Hong Kong SAR\\
    \textsuperscript{\rm 4}University of Chinese Academy of Sciences, Beijing, China\\
    \textsuperscript{\rm 5}Institute of Automation, Chinese Academy of Sciences, China\\
    \{chenxi.liu,mingzhao.li,hongyuan.zhang,gaofeng.meng\}@cair-cas.org.hk,\\
    \{liuyc27, chenziyi\}@mail.sysu.edu.cn, hao.miao@polyu.edu.hk
}

\begin{document}

\maketitle

\begin{abstract}
Automated seizure detection from electroencephalography (EEG) is essential for continuous neurological monitoring, particularly for subclinical epileptic seizures that may exhibit only subtle electrographic changes. Existing time-series methods are often designed for fixed EEG channel configurations, thereby limiting their applicability to heterogeneous EEG recordings with irregular channel layouts. Although visual and language modeling offer promising alternatives, aligning heterogeneous EEG representations with clinical semantics remains challenging. We introduce \textbf{ViTexSZ}, a heterogeneous \textbf{V}ision–\textbf{T}ext knowledge distillation framework for EEG \textbf{S}ei\textbf{Z}ure detection. ViTexSZ converts EEG recordings into structured waveform images and introduces a query-based multi-channel alignment module that maps source-dependent visual features into a unified token space. A heterogeneous teacher further integrates the aligned EEG representations with clinical prompts through a multimodal large language model, associating high-level clinical semantics with seizure-related evidence. Vision–text knowledge distillation then transfers the teacher representations to a lightweight student during detection. Experiments on four EEG seizure datasets demonstrate the generalizability of ViTexSZ across both subclinical and general seizure detection scenarios, achieving the highest accuracy on all datasets and relative improvements of up to 12.9\% over the second-best baselines, demonstrating its effectiveness.
\end{abstract}


\section{Introduction}

Recent advances in artificial intelligence have enabled the modeling of electroencephalography (EEG) signals, positioning automated seizure detection as a central problem in medical time-series analysis~\cite{wang2026brastorm}. Timely identification of seizure events supports continuous neurological monitoring and clinical decision-making~\cite{abou2022noninvasive}. Such needs are particularly evident in subclinical epileptic seizures, which may manifest only through subtle electrographic changes and therefore remain unnoticed during routine monitoring~\cite{he2025eeg}. Reliable detection of seizure events could facilitate earlier recognition of abnormal brain activity and provide evidence for subsequent clinical assessment and intervention~\cite{tang2025eeg}.

Existing EEG seizure analysis studies capture cross-channel dependencies through adaptive graphs~\cite{fan2025medgnn} or synchronization-aware aggregation~\cite{yu2026tech}, and multi-scale temporal dependencies via attention-based modeling~\cite{affes2022personalized}. However, their effectiveness is often tied to fixed channel configurations, limiting the applicability to heterogeneous EEG settings. 
Recently, vision~\cite{zhou2025enhancing} and language models~\cite{wanglong} provide a complementary direction by transforming EEG into structured images or incorporating textual knowledge into representation learning~\cite{hossain2019applying,sun2025automated,riazi2026szxai,zhang2025uwt}. 
These studies demonstrate that image and language provide more comprehensive knowledge for EEG physiological signals, establishing visual and language modeling as a promising direction for seizure detection. However, effectively exploiting knowledge across heterogeneous EEG sources while maintaining efficient seizure detection remains challenging.

\begin{figure}[t]
    \centering
    \includegraphics[width=1\linewidth]{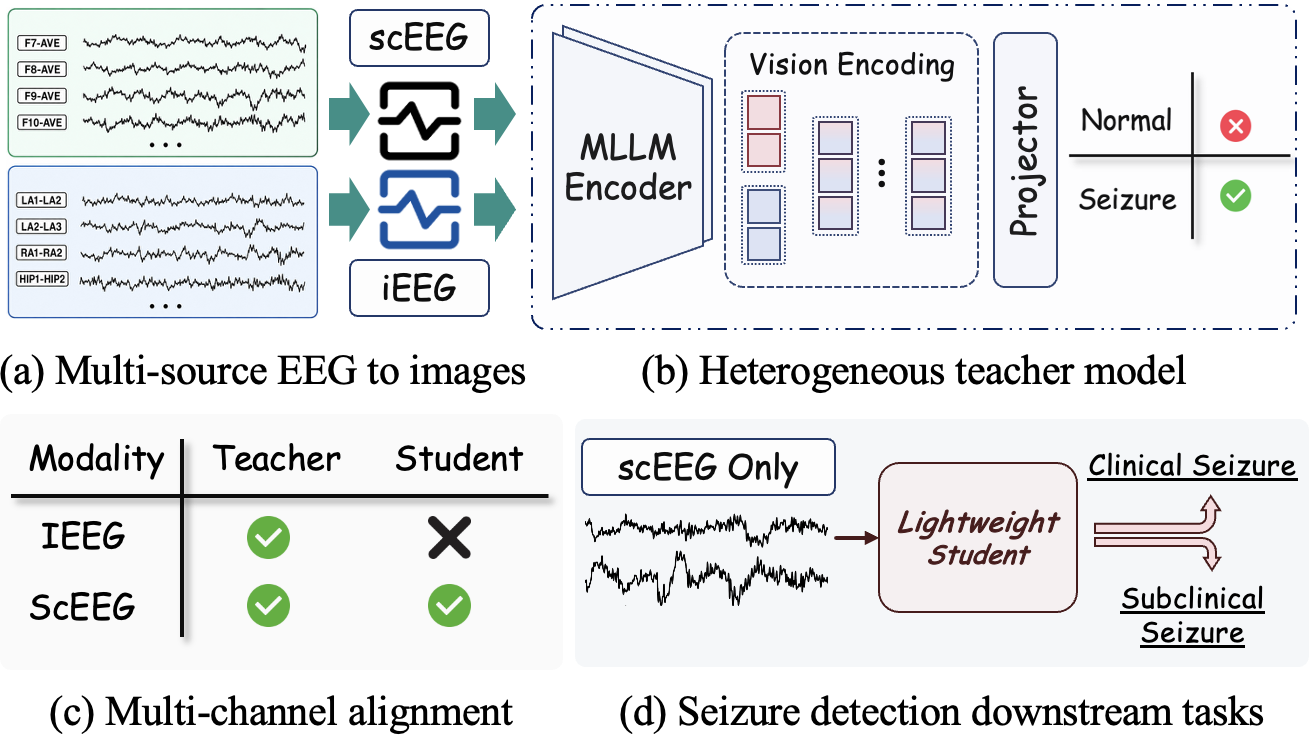}
    \caption{ViTexSZ training and usage.}
    \label{fig:Motivation}
\end{figure}

The first challenge is how to transfer seizure knowledge across heterogeneous EEG acquisition settings. Due to variations in clinical protocols and equipment, EEG recordings often exhibit diverse channel configurations and spatial coverage, such as high-density intracranial EEG and standard scalp EEG recordings with different channels~\cite{gu2026cerebragloss,hogan2025scaling}, as shown in Figure~\ref{fig:Motivation} (a). These multi-source discrepancies lead to inconsistent visual structures and feature distributions, limiting the generalization of standard architectures across different channel setups without costly re-training. The second challenge is how to bridge the multi-modal semantic gap between high-level clinical language and fine-grained EEG dynamics. Clinical texts, such as Electronic Health Records~\cite{11461333}, describe diagnostic insights and seizure characteristics at an abstract semantic level, whereas EEG representations encode waveform morphology and multi-channel temporal dynamics.

To address these challenges, we propose ViTexSZ, a heterogeneous vision--text knowledge distillation framework for EEG seizure detection. As illustrated in Figure~\ref{fig:Motivation} (a), ViTexSZ first converts EEG signals into waveform images and employs a query-based multi-channel alignment module to project source-dependent visual features into a unified token space. A heterogeneous teacher, shown in Figure~\ref{fig:Motivation} (b), then integrates EEG visual representations with clinical prompts through a multimodal large language model (MLLM), thereby aligning seizure-related visual patterns with high-level clinical semantics. Next, token-level vision--text knowledge distillation transfers the teacher's multimodal representations to a lightweight student that requires only scEEG images and clinical prompts during inference, as illustrated in Figure~\ref{fig:Motivation} (c). Finally, as shown in Figure~\ref{fig:Motivation} (d), ViTexSZ enables efficient scEEG-based detection of both subclinical seizures and clinically observable seizures.


Our main contributions are summarized as follows:
\begin{itemize}
\item We propose \textbf{ViTexSZ}, a \textbf{V}ision--\textbf{T}ext knowledge distillation \textbf{S}ei\textbf{Z}ure detection framework that transfers the augmented representation of heterogeneous EEG and EHR while retaining lightweight detection.
\item We design a multi-channel alignment module that maps source-specific visual representations into a unified token space, enabling knowledge transfer across varying channel configurations.
\item We develop a heterogeneous teacher that grounds clinical semantics in seizure-related visual patterns and aligns the resulting representations to a lightweight student through token-level vision--text knowledge distillation.
\item Extensive experiments on four EEG seizure datasets demonstrate the generalizability and effectiveness of ViTexSZ across both subclinical and seizure detection tasks, achieving relative improvements of 12.9\% and 7.9\% over the second-best baselines in sensitivity and F1 score.
\end{itemize}

\section{Related Work}

\textbf{Time series-centric methods for EEG analysis}.
Existing medical time-series methods typically model EEG signals from cross-channel and multi-scale temporal perspectives.
Cross-channel methods characterize interactions among physiological channels. 
For instance, MedGNN~\cite{fan2025medgnn} constructs adaptive graphs at multiple resolutions, and TeCh~\cite{yu2026tech} models global synchronization patterns through centralized token aggregation. The spatial structures learned by these methods are are often coupled to specific channel configurations, making them sensitive to montage~\cite{gu2026cerebragloss} variation and missing channels. Multi-scale temporal methods focus on seizure evolution across different temporal resolutions, such as Transformer-based methods, which combine the local waveform morphology with self-attention for long-range sequence modeling and time-step-level detection~\cite{affes2022personalized}. Nevertheless, the combination of hierarchical encoding and long-sequence attention introduces considerable computational overhead.

\noindent \textbf{Visual and language-augmented EEG modeling}. Recent advances in visual representation learning have motivated the transformation of EEG into structured visual representations~\cite{DBLP:conf/iclr/CarzanigaHHSR25}. Visual approaches encode waveform morphology and inter-channel organization through EEG images, thereby enabling the transfer of visual priors to downstream time-series tasks~\cite{hossain2019applying,sun2025automated}. For example, VIPEEGNet~\cite{sun2025automated} employs an EEG-to-image transformation together with a pretrained visual backbone for brain-activity classification. Language-augmented approaches associate EEG with clinical concepts or task prompts through semantic alignment~\cite{riazi2026szxai,gu2026cerebragloss,ye2025medualtime}. These studies demonstrate that image and language provide more comprehensive knowledge for physiological signals, establishing visual and language modeling as a promising direction for EEG representation learning.

\section{Preliminaries}

\textbf{Definition 1 (Heterogeneous EEG Signals).} Let $\mathcal{D}=\{(\mathbf{X}_{s}^{n}, \mathbf{X}_{i}^{n}, y^{n})\}_{n=1}^{N}$ denote a dataset containing $N$ EEG segments. Each segment is a temporally continuous multichannel time series extracted from an EEG recording, where $\mathbf{X}_{s}^{n}\in\mathbb{R}^{C_s\times T}$ and $\mathbf{X}_{i}^{n}\in\mathbb{R}^{C_i\times T}$ represent scalp EEG (scEEG) and intracranial EEG (iEEG), respectively. Here, $C_s$ and $C_i$ denote the numbers of scEEG and iEEG channels, $T$ is the number of time steps, and $y^{n}$ is the seizure label of the $n$-th segment. The two EEG modalities are not paired for every data source; that is, either $\mathbf{X}_{s}^{n}$ or $\mathbf{X}_{i}^{n}$ may be unavailable, while at least one modality is observed.

\begin{figure}[t]
    \centering
    \includegraphics[width=0.96\linewidth]{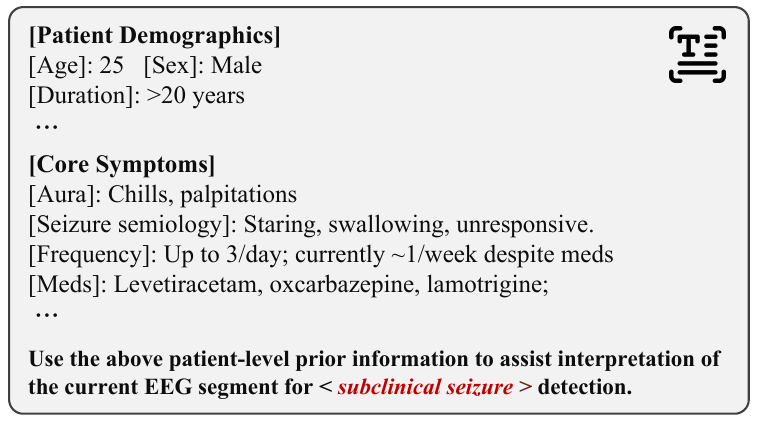}
    \caption{An Example of a patient-level clinical prompt.}
    \label{fig:prompt_example}
\end{figure}

\noindent \textbf{Definition 2 (Clinical Prompts).} A clinical prompt $\mathcal{P}$ is a textual sequence that integrates the subset of Electronic Health Records and detection instructions, including the definitions and characteristics of different seizure events, as shown in Figure~\ref{fig:prompt_example}.

\begin{figure}[h]
    \centering
    \hspace*{-2.5pt}%
    \includegraphics[width=\linewidth,trim=0 0 0 0,clip]{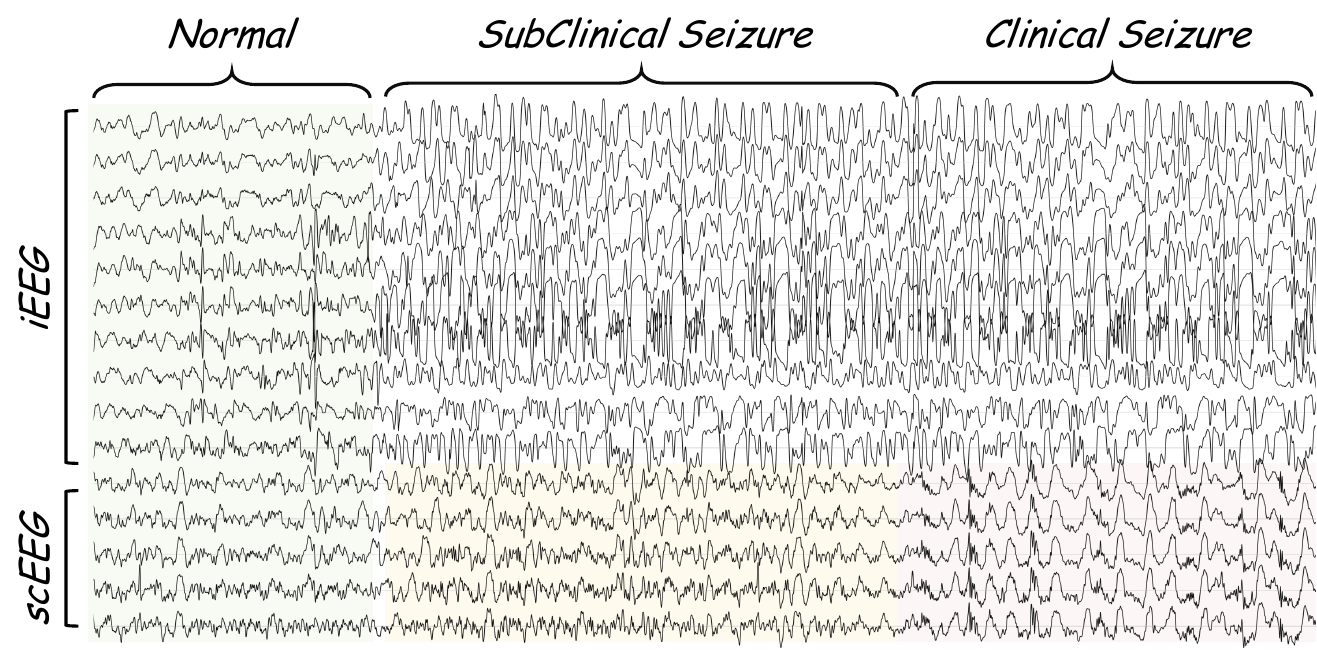}
    \caption{Ictal signatures are clearest on iEEG, partially visible on scEEG in CS, and most subtle in SCS.}
    \label{fig:seizure_types}
\end{figure}

\noindent \textbf{Definition 3 (Seizure Events).}
For the $n$-th EEG segment, we define the binary seizure label $y^{n}\in\{0,1\}$ and, for seizure segments, the event subtype $c^{n}\in\{\mathrm{SCS},\mathrm{CS}\}$:
\begin{equation}
y^{n}
=
\mathbb{I}
\left(
c^{n}\in\{\mathrm{SCS},\mathrm{CS}\}
\right),
\end{equation}
where $y^{n}=0$ denotes normal activity and $y^{n}=1$ denotes a seizure event. A subclinical seizure (SCS) exhibits electrographic ictal activity without observable clinical manifestations, whereas a clinical seizure (CS) is accompanied by overt behavioral or physiological signs~\cite{he2025eeg}. Compared with CS, SCS is substantially more challenging to detect because it lacks external clinical cues and may present only subtle in scalp EEG, as shown in Figure~\ref{fig:seizure_types}.


\paragraph{Problem Definition.} We study seizure detection from multivariate scEEG $\mathbf{X}_{s}^{n}$, where paired iEEG $\mathbf{X}_{i}^{n}$, when available; and a clinical prompt $\mathcal{P}$ is a textual sequence that integrates task instructions with clinically relevant contexts. The teacher model $f_T\in\mathcal{F}$ predicts the seizure state as $\hat{y}_{T}^{n}=f_T(\mathbf{X}_{s}^{n},\mathbf{X}_{i}^{n},\mathcal{P})$. The objective is to transfer its knowledge to a lightweight detector $f_S\in\mathcal{F}$, which predicts the seizure state during inference as $\hat{y}^{n}_S=f_S(\mathbf{X}_{s}^{n},\mathcal{P})$.

\section{Methodology}

\subsection{Overall Framework}

\begin{figure*}[t]
\centering
\includegraphics[width=\linewidth]{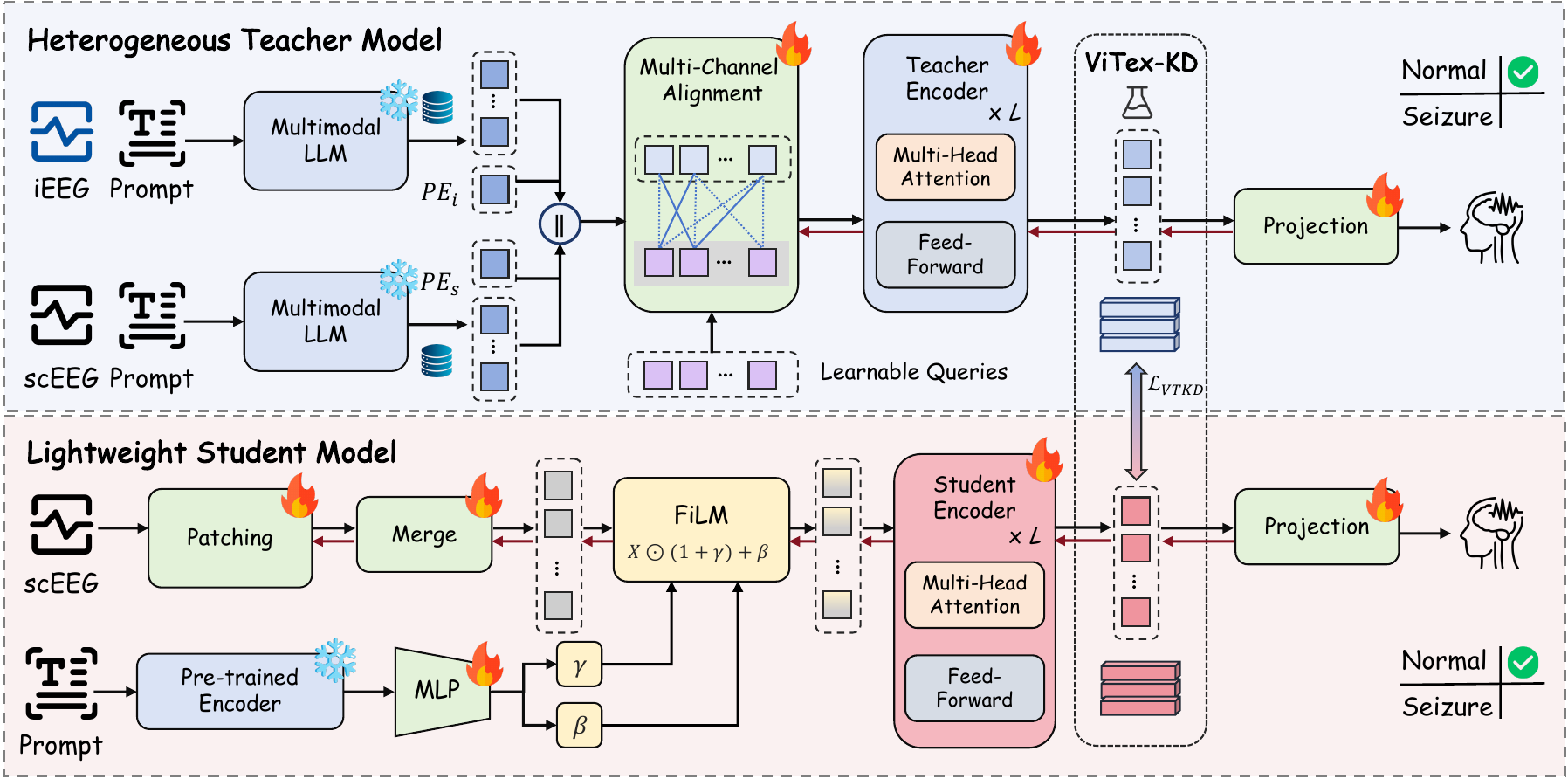}
\caption{Overall framework of ViTexSZ.}
\label{fig:framework}
\end{figure*}

ViTexSZ consists of a heterogeneous teacher model, a vision-text knowledge distillation, and a lightweight student model, as illustrated in Fig.~\ref{fig:framework}.

\textbf{Heterogeneous Teacher Model} encodes EEG images and clinical prompts using a \textit{multimodal large language model (MLLM)}, a \textit{multi-channel alignment} module, and a \textit{vision-text encoder} module to generate high-quality vision-text representations.

\textbf{Vision-Text Knowledge Distillation} transfers the heterogeneous teacher's representations to the student, with iEEG information, MLLM-derived knowledge, or both serving as privileged information available only during training.

\textbf{Lightweight Student Model} consists of a lightweight \textit{pre-trained encoder}, a \textit{feature-wise linear modulation (FiLM)} module, and a \textit{vision encoder}. The student learns from the teacher's privileged representations and enables efficient seizure detection.

\subsection{Heterogeneous Teacher Model}

\subsubsection{Multimodal LLM Encoding}
Multichannel EEG images encode temporal dynamics and inter-channel organization in a unified visual representation, enabling MLLMs to emulate clinicians' visual inspection of EEG traces~\cite{sun2025automated,gu2026cerebragloss}. Therefore, we first convert each available EEG time series $\mathbf{X}_{m}$ into an image $\mathbf{I}_{m}\in\mathbb{R}^{H\times W\times C}$, where $m\in\{s,i\}$ denotes the scEEG or iEEG. Then, the heterogeneous teacher employs a frozen multimodal LLM as a feature extractor to jointly encode $\mathbf{I}_{m}$ and $\mathbf{P}$:
\begin{equation}
\mathbf{H}_{m}
=
\mathcal{F}_{\mathrm{MLLM}}^{(\ell)}
\left(
\left[
\mathcal{G}
\bigl(\mathcal{E}_{\mathrm{V}}(\mathbf{I}_{m})\bigr);
\mathcal{E}_{\mathrm{T}}(\mathcal{P})
\right];
\boldsymbol{\theta}^{*}
\right),
\end{equation}
where $\mathbf{H}_{m}\in\mathbb{R}^{L_m\times D_0}$ denotes the $\ell$-th-layer hidden states of the frozen MLLM, and $\mathcal{E}{\mathrm{V}}$, $\mathcal{E}_{\mathrm{T}}$, and $\mathcal{G}$ are the visual encoder, text encoder, and projector, respectively.

To preserve the source identity of the two EEG modalities, we introduce learnable positional embeddings $\mathbf{PE}_{s},\mathbf{PE}_{i}\in\mathbb{R}^{D_0}$. The modality-aware features are projected into a shared $D$-dimensional space:
\begin{equation}
\widetilde{\mathbf{H}}_{m}
=
\left(
\mathbf{H}_{m}
+
\mathbf{PE}_{m}
\right)
\mathbf{W}_{a}
+
\mathbf{b}_{a},
\end{equation}
where $m\in\{s,i\}$ denotes the scEEG or iEEG, and
$\mathbf{W}_{a}\in\mathbb{R}^{D_0\times D}$ and
$\mathbf{b}_{a}\in\mathbb{R}^{D}$ are trainable projection parameters.

\subsubsection{Multi-Channel Alignment}
The EEG modalities differ in channel configuration, ranging from \textit{scalp electrodes under the international 10--20 system} to \textit{patient-specific intracranial electrodes}~\cite{DBLP:conf/iclr/CarzanigaHHSR25}, which complicates efficient cross-modal processing. We therefore introduce learnable queries $\mathbf{Q}\in\mathbb{R}^{N_q\times D}$ to aggregate each available modality into $N_q$ aligned tokens. Three trainable projections $\psi_q$, $\psi_k$, and $\psi_v$ then map the queries and EEG tokens into compact representations, from which the query-to-token similarity matrix is computed as:
\begin{equation}
\mathbf{M}_{m}
=
\mathcal{F}_{Softmax}
\left(
\psi_q(\mathbf{Q})
\psi_k(\widetilde{\mathbf{H}}_{m})^{\top}
\right),
\end{equation}
where $\mathbf{M}_{m}\in\mathbb{R}^{N_q\times L_m}$.

Based on $\mathbf{M}_{m}$, seizure-relevant information is retrieved from each EEG modality and then fused into a fixed-length aligned sequence:
\begin{equation}
\mathbf{A}_m
=
\omega_a
\left(
\mathop{\mathcal{F}_{Concat}}
\left(\mathbf{M}_{m}
\psi_v(\widetilde{\mathbf{H}}_{m})\right)
\right)
\oplus
\mathbf{Q},
\end{equation}
where $\mathbf{A}_m\in\mathbb{R}^{N_q\times D}$, $\omega_a$ denotes a trainable projection, and $\oplus$ denotes residual addition.

\subsubsection{Teacher Encoder}
To further capture seizure task-specific dependencies among the aligned visual-text tokens, we employ a trainable vision encoder. Specifically, $\mathbf{A}_{m}$ is refined by an encoder consisting of $L$ Transformer layers to capture task-specific interactions and enhance seizure-discriminative representations. We initialize $\mathbf{Z}_{m}^{T,(0)}=\mathbf{A}_{m}$. The $r$-th Transformer layer is formulated as:
\begin{equation}
\begin{aligned}
\mathbf{U}_{m}^{T,(r)}
&=
\mathbf{Z}_{m}^{T,(r-1)}
\oplus
\mathcal{F}_{\mathrm{MSA}}^{(r)}
\left(
\mathcal{F}_{\mathrm{LN}}
\left(
\mathbf{Z}_{m}^{T,(r-1)}
\right)
\right),\\
\mathbf{Z}_{m}^{T,(r)}
&=
\mathbf{U}_{m}^{T,(r)}
\oplus
\mathcal{F}_{\mathrm{FFN}}^{(r)}
\left(
\mathcal{F}_{\mathrm{LN}}
\left(
\mathbf{U}_{m}^{T,(r)}
\right)
\right),
\end{aligned}
\end{equation}
where $r=1,\ldots,L$. The final teacher representation is denoted by
\begin{equation}
\mathbf{Z}_{m}^{T}
=
\mathbf{Z}_{m}^{T,(L)},
\qquad
\mathbf{Z}_{m}^{T}\in\mathbb{R}^{N_q\times D}.
\end{equation}

A trainable projection head directly maps the teacher tokens to the seizure
prediction:
\begin{equation}
\hat{\mathbf{y}}_{m}^{T}
=
\mathcal{F}_{\mathrm{Softmax}}
\left(
g_T\left(\mathbf{Z}_{m}^{T}\right)
\right),
\qquad
\mathcal{L}_{\mathrm{DET}}^{T}
=
\mathcal{F}_{\mathrm{CE}}
\left(
\hat{\mathbf{y}}_{m}^{T},y
\right).
\end{equation}

\subsection{Vision-Text Knowledge Distillation}

Vision-text knowledge distillation transfers seizure-relevant
representations available only during training from the heterogeneous teacher to the lightweight student. The teacher exploits MLLM-derived visual-text knowledge and, when available, iEEG information, whereas the student relies only on scEEG and clinical prompts. This allows the student to benefit from complementary privileged information without introducing additional inference cost.

For each available EEG modality, we align the complete teacher and student
token matrices using a normalized Smooth L1 objective:
\begin{equation}
\mathcal{L}_{\mathrm{VTKD}}
=
\frac{1}{|\mathcal{M}|}
\sum_{m\in\mathcal{M}}
\frac{1}{N_qD}
\left\|
\mathbf{Z}^{S}
-
\mathrm{sg}
\left[
\mathbf{Z}_{m}^{T}
\right]
\right\|_{\mathrm{SmoothL1}},
\end{equation}
where $\mathrm{sg}[\cdot]$ denotes the stop-gradient operation and
$\|\cdot\|_{\mathrm{SmoothL1}}$ applies the Smooth L1 penalty element-wise over the aligned token matrix. 

The knowledge distillation optimization is formulated as
\begin{equation}
\begin{aligned}
\boldsymbol{\theta}_{T}^{*}
&=
\arg\min_{\boldsymbol{\theta}_{T}}
\mathcal{L}_{\mathrm{DET}}^{T},
\\
\boldsymbol{\theta}_{S}^{*}
&=
\arg\min_{\boldsymbol{\theta}_{S}}
\left[
\mathcal{L}_{\mathrm{DET}}^{S}
+
\lambda_{\mathrm{VTKD}}
\mathcal{L}_{\mathrm{VTKD}}
\left(
\boldsymbol{\theta}_{S};
\mathrm{sg}
\left[
\boldsymbol{\theta}_{T}^{*}
\right]
\right)
\right],
\end{aligned}
\end{equation}
where $\lambda_{\mathrm{VTKD}}$ balances seizure detection and vision-text
knowledge distillation. The optimized teacher parameters
$\boldsymbol{\theta}_{T}^{*}$ remain frozen during student training.

During cross-patient inference, the heterogeneous teacher branch is removed, and only the lightweight student is retained to predict $\hat{\mathbf{y}}^{S}$ from scEEG and the prompt.

\subsection{Lightweight Student Model}

The lightweight student is designed for efficient and non-invasive seizure detection using scEEG as the only physiological input. In addition to visual EEG representations, it incorporates prompt-conditioned feature modulation to inject clinical semantic priors into the student tokens with limited computational overhead.

\subsubsection{scEEG Patching}

Given the scEEG image $\mathbf{I}_{s}$, we first partition it into non-overlapping patches and project them into a sequence of visual tokens:
\begin{equation}
\mathbf{X}_{p}^{S}
=
\mathcal{F}_{\mathrm{Patch}}
\left(
\mathbf{I}_{s}
\right),
\end{equation}
the trainable merging operator then aggregates neighboring patches and produces $N_q$ student tokens:
\begin{equation}
\mathbf{X}^{S}
=
\mathcal{F}_{\mathrm{Merge}}
\left(
\mathbf{X}_{p}^{S}
\right),
\end{equation}
where $\mathbf{X}_{p}^{S}\in\mathbb{R}^{N_p\times D}$ and $\mathbf{X}^{S}\in\mathbb{R}^{N_q\times D}$.
The token number $N_q$ matches that of the teacher representation, enabling direct token-level knowledge transfer.

\subsubsection{Prompt-Conditioned FiLM}

$\mathcal{P}$ is encoded by a pre-trained text encoder, and a trainable multilayer perceptron generates channel-wise modulation parameters:
\begin{equation}
\mathbf{p}
=
\mathcal{E}_{\mathrm{P}}
\left(
\mathcal{P};
\boldsymbol{\phi}^{*}
\right),
\qquad
\left[
\boldsymbol{\gamma};
\boldsymbol{\beta}
\right]
=
\mathcal{F}_{\mathrm{MLP}}
\left(
\mathbf{p}
\right),
\end{equation}
where $\mathbf{p}\in\mathbb{R}^{D_p}$ and $\boldsymbol{\gamma},\boldsymbol{\beta}\in\mathbb{R}^{D}$.

The prompt-conditioned student tokens are obtained through feature-wise linear modulation:
\begin{equation}
\widetilde{\mathbf{X}}^{S}
=
\mathbf{X}^{S}
\odot
\left(
\mathbf{1}
+
\boldsymbol{\gamma}
\right)
\oplus
\boldsymbol{\beta},
\end{equation}
where $\boldsymbol{\gamma}$ and $\boldsymbol{\beta}$ are broadcast across the $N_q$ tokens. This operation adaptively modulates seizure-relevant feature dimensions according to the clinical prompt.

\subsubsection{Student Encoder and Detection}

The modulated tokens are refined by a lightweight student encoder consisting of $L_S$ Transformer layers. We initialize $\mathbf{Z}^{S,(0)}=\widetilde{\mathbf{X}}^{S}$. The $r$-th student layer is formulated as
\begin{equation}
\begin{aligned}
\mathbf{U}^{S,(r)}
&=
\mathbf{Z}^{S,(r-1)}
\oplus
\mathcal{F}_{\mathrm{MSA},S}^{(r)}
\left(
\mathcal{F}_{\mathrm{LN}}
\left(
\mathbf{Z}^{S,(r-1)}
\right)
\right),\\
\mathbf{Z}^{S,(r)}
&=
\mathbf{U}^{S,(r)}
\oplus
\mathcal{F}_{\mathrm{FFN},S}^{(r)}
\left(
\mathcal{F}_{\mathrm{LN}}
\left(
\mathbf{U}^{S,(r)}
\right)
\right),
\end{aligned}
\end{equation}
where $r=1,\ldots,L_S$ and the final student representation $\mathbf{Z}^{S}\in\mathbb{R}^{N_q\times D}$ is denoted by
\begin{equation}
\mathbf{Z}^{S}
=
\mathbf{Z}^{S,(L_S)}.
\end{equation}

The token representation $\mathbf{Z}^{S}$ is directly supervised by $\mathcal{L}_{\mathrm{VTKD}}$ and mapped to the seizure prediction through a trainable projection head:
\begin{equation}
\hat{\mathbf{y}}^{S}
=
\mathcal{F}_{\mathrm{Softmax}}
\left(
g_S
\left(
\mathbf{Z}^{S}
\right)
\right),
\qquad
\mathcal{L}_{\mathrm{DET}}^{S}
=
\mathcal{F}_{\mathrm{CE}}
\left(
\hat{\mathbf{y}}^{S},
y
\right).
\end{equation}


\begin{table*}[t]
  \centering
  \fontsize{7}{8}\selectfont
  \setlength{\tabcolsep}{2pt}
  \renewcommand{\arraystretch}{0.93}
  \begin{tabular*}{\textwidth}{@{\extracolsep{\fill}}
    >{\centering\arraybackslash}m{1.4cm}c*{8}{c}@{}}
    \toprule
    &
    & \multicolumn{4}{c}{\textbf{Dual-SCS}}
    & \multicolumn{4}{c}{\textbf{Dual-CS}} \\
    \cmidrule(lr){3-6} \cmidrule(lr){7-10}
    \textbf{Type} & \textbf{Method}
    & \textbf{Accuracy}
    & \textbf{F1 Score}
    & \textbf{Sensitivity}
    & \textbf{AUC}
    & \textbf{Accuracy}
    & \textbf{F1 Score}
    & \textbf{Sensitivity}
    & \textbf{AUC} \\
    \midrule
    \sidegroup{7}{Time-Series}
    & TCN & 0.701\,$\pm$\,0.104 & 0.630\,$\pm$\,0.160 & 0.550\,$\pm$\,0.217 & 0.754\,$\pm$\,0.113 & 0.710\,$\pm$\,0.045 & 0.691\,$\pm$\,0.051 & 0.634\,$\pm$\,0.081 & 0.769\,$\pm$\,0.080 \\
    & Transformer & 0.733\,$\pm$\,0.065 & 0.754\,$\pm$\,0.039 & 0.793\,$\pm$\,0.048 & 0.787\,$\pm$\,0.071 & 0.698\,$\pm$\,0.046 & 0.688\,$\pm$\,0.057 & 0.664\,$\pm$\,0.153 & 0.736\,$\pm$\,0.081 \\
    & MedGNN & 0.578\,$\pm$\,0.016 & 0.545\,$\pm$\,0.051 & 0.509\,$\pm$\,0.116 & 0.573\,$\pm$\,0.033 & 0.530\,$\pm$\,0.041 & 0.581\,$\pm$\,0.184 & \third{0.737\,$\pm$\,0.121} & 0.498\,$\pm$\,0.053 \\
    & TimeMIL & 0.680\,$\pm$\,0.019 & 0.688\,$\pm$\,0.029 & 0.702\,$\pm$\,0.087 & 0.728\,$\pm$\,0.017 & 0.692\,$\pm$\,0.036 & 0.679\,$\pm$\,0.024 & 0.633\,$\pm$\,0.038 & 0.713\,$\pm$\,0.055 \\
    & TeCh & 0.727\,$\pm$\,0.044 & 0.733\,$\pm$\,0.052 & 0.749\,$\pm$\,0.105 & 0.791\,$\pm$\,0.073 & 0.703\,$\pm$\,0.040 & 0.700\,$\pm$\,0.071 & 0.694\,$\pm$\,0.125 & 0.745\,$\pm$\,0.060 \\
    & ConvNeXt-Seizure & 0.779\,$\pm$\,0.018 & 0.769\,$\pm$\,0.019 & 0.731\,$\pm$\,0.076 & 0.842\,$\pm$\,0.056 & 0.680\,$\pm$\,0.031 & 0.688\,$\pm$\,0.062 & 0.716\,$\pm$\,0.123 & 0.697\,$\pm$\,0.098 \\
    \midrule
    \sidegroup{6}{Visual / V-L}
    & BLIP & 0.814\,$\pm$\,0.018 & 0.812\,$\pm$\,0.042 & 0.819\,$\pm$\,0.132 & \third{0.890\,$\pm$\,0.015} & 0.686\,$\pm$\,0.054 & 0.702\,$\pm$\,0.092 & \second{0.752\,$\pm$\,0.134} & 0.722\,$\pm$\,0.081 \\
    & CLIP & 0.707\,$\pm$\,0.097 & 0.701\,$\pm$\,0.116 & 0.724\,$\pm$\,0.229 & 0.852\,$\pm$\,0.028 & 0.582\,$\pm$\,0.048 & 0.522\,$\pm$\,0.147 & 0.596\,$\pm$\,0.208 & 0.689\,$\pm$\,0.076 \\
    & ResNet-18 & \third{0.826\,$\pm$\,0.036} & \second{0.846\,$\pm$\,0.024} & \best{0.938\,$\pm$\,0.056} & 0.865\,$\pm$\,0.035 & 0.717\,$\pm$\,0.065 & 0.712\,$\pm$\,0.052 & 0.678\,$\pm$\,0.059 & 0.752\,$\pm$\,0.077 \\
    & ConvNeXt v2 & 0.801\,$\pm$\,0.056 & 0.780\,$\pm$\,0.074 & 0.712\,$\pm$\,0.113 & 0.873\,$\pm$\,0.056 & 0.675\,$\pm$\,0.049 & 0.614\,$\pm$\,0.137 & 0.558\,$\pm$\,0.140 & 0.750\,$\pm$\,0.092 \\
    & SigLIP & \second{0.827\,$\pm$\,0.072} & \third{0.841\,$\pm$\,0.064} & \second{0.891\,$\pm$\,0.073} & \second{0.908\,$\pm$\,0.053} & \third{0.721\,$\pm$\,0.065} & \third{0.716\,$\pm$\,0.108} & 0.727\,$\pm$\,0.104 & \third{0.791\,$\pm$\,0.087} \\
    & VIPEEGNet & 0.741\,$\pm$\,0.063 & 0.726\,$\pm$\,0.065 & 0.678\,$\pm$\,0.091 & 0.819\,$\pm$\,0.074 & \second{0.755\,$\pm$\,0.038} & \second{0.730\,$\pm$\,0.053} & 0.659\,$\pm$\,0.076 & \best{0.840\,$\pm$\,0.047} \\
    \midrule
    \sidegroup{1}{Ours}
    & \textbf{ViTexSZ} & \best{0.851\,$\pm$\,0.029} & \best{0.849\,$\pm$\,0.039} & \third{0.848\,$\pm$\,0.102} & \best{0.929\,$\pm$\,0.022} & \best{0.766\,$\pm$\,0.033} & \best{0.788\,$\pm$\,0.020} & \best{0.849\,$\pm$\,0.057} & \second{0.802\,$\pm$\,0.051} \\
    \addlinespace[2pt]
    \midrule[\lightrulewidth]
    \midrule[\lightrulewidth]
    \addlinespace[2pt]
    &
    & \multicolumn{4}{c}{\textbf{CHB-MIT}}
    & \multicolumn{4}{c}{\textbf{TUSZ}} \\
    \cmidrule(lr){3-6} \cmidrule(lr){7-10}
    \textbf{Type} & \textbf{Method}
    & \textbf{Accuracy}
    & \textbf{F1 Score}
    & \textbf{Sensitivity}
    & \textbf{AUC}
    & \textbf{Accuracy}
    & \textbf{F1 Score}
    & \textbf{Sensitivity}
    & \textbf{AUC} \\
    \midrule
    \sidegroup{7}{Time-Series}
    & TCN & 0.687\,$\pm$\,0.025 & 0.656\,$\pm$\,0.072 & 0.613\,$\pm$\,0.145 & 0.732\,$\pm$\,0.052 & 0.695\,$\pm$\,0.036 & 0.705\,$\pm$\,0.038 & 0.654\,$\pm$\,0.060 & 0.757\,$\pm$\,0.048 \\
    & Transformer & 0.679\,$\pm$\,0.029 & 0.658\,$\pm$\,0.066 & 0.626\,$\pm$\,0.118 & 0.742\,$\pm$\,0.037 & 0.703\,$\pm$\,0.028 & \third{0.733\,$\pm$\,0.024} & 0.734\,$\pm$\,0.053 & 0.761\,$\pm$\,0.042 \\
    & MedGNN & 0.546\,$\pm$\,0.012 & 0.523\,$\pm$\,0.081 & 0.512\,$\pm$\,0.141 & 0.542\,$\pm$\,0.012 & 0.702\,$\pm$\,0.028 & 0.726\,$\pm$\,0.019 & 0.705\,$\pm$\,0.019 & 0.758\,$\pm$\,0.043 \\
    & TimeMIL & 0.715\,$\pm$\,0.064 & 0.696\,$\pm$\,0.084 & 0.656\,$\pm$\,0.116 & 0.765\,$\pm$\,0.087 & 0.710\,$\pm$\,0.041 & 0.725\,$\pm$\,0.043 & 0.688\,$\pm$\,0.050 & \third{0.774\,$\pm$\,0.056} \\
    & TeCh & \third{0.719\,$\pm$\,0.026} & \third{0.711\,$\pm$\,0.021} & \third{0.680\,$\pm$\,0.050} & \third{0.777\,$\pm$\,0.037} & 0.660\,$\pm$\,0.015 & 0.707\,$\pm$\,0.042 & \second{0.747\,$\pm$\,0.116} & 0.695\,$\pm$\,0.027 \\
    & ConvNeXt-Seizure & \second{0.743\,$\pm$\,0.054} & \second{0.747\,$\pm$\,0.043} & \second{0.747\,$\pm$\,0.064} & \best{0.804\,$\pm$\,0.070} & \second{0.730\,$\pm$\,0.027} & \best{0.749\,$\pm$\,0.017} & 0.720\,$\pm$\,0.053 & \second{0.798\,$\pm$\,0.028} \\
    \midrule
    \sidegroup{6}{Visual / V-L}
    & BLIP & 0.644\,$\pm$\,0.042 & 0.665\,$\pm$\,0.023 & 0.638\,$\pm$\,0.127 & 0.678\,$\pm$\,0.057 & 0.677\,$\pm$\,0.065 & 0.701\,$\pm$\,0.072 & \third{0.747\,$\pm$\,0.064} & 0.726\,$\pm$\,0.071 \\
    & CLIP & 0.636\,$\pm$\,0.032 & 0.650\,$\pm$\,0.079 & 0.634\,$\pm$\,0.140 & 0.713\,$\pm$\,0.092 & 0.634\,$\pm$\,0.045 & 0.682\,$\pm$\,0.048 & \best{0.805\,$\pm$\,0.076} & 0.690\,$\pm$\,0.031 \\
    & ResNet-18 & 0.669\,$\pm$\,0.066 & 0.655\,$\pm$\,0.034 & 0.625\,$\pm$\,0.106 & 0.713\,$\pm$\,0.078 & 0.709\,$\pm$\,0.026 & 0.728\,$\pm$\,0.028 & 0.702\,$\pm$\,0.072 & 0.770\,$\pm$\,0.038 \\
    & ConvNeXt v2 & 0.633\,$\pm$\,0.031 & 0.627\,$\pm$\,0.063 & 0.630\,$\pm$\,0.148 & 0.684\,$\pm$\,0.080 & 0.664\,$\pm$\,0.035 & 0.690\,$\pm$\,0.023 & 0.669\,$\pm$\,0.039 & 0.720\,$\pm$\,0.046 \\
    & SigLIP & 0.649\,$\pm$\,0.071 & 0.635\,$\pm$\,0.102 & 0.624\,$\pm$\,0.151 & 0.680\,$\pm$\,0.086 & 0.686\,$\pm$\,0.028 & 0.699\,$\pm$\,0.021 & 0.654\,$\pm$\,0.043 & 0.759\,$\pm$\,0.035 \\
    & VIPEEGNet & 0.709\,$\pm$\,0.082 & 0.673\,$\pm$\,0.111 & 0.612\,$\pm$\,0.141 & 0.767\,$\pm$\,0.081 & \third{0.714\,$\pm$\,0.031} & 0.731\,$\pm$\,0.028 & 0.698\,$\pm$\,0.029 & 0.774\,$\pm$\,0.038 \\
    \midrule
    \sidegroup{1}{Ours}
    & \textbf{ViTexSZ} & \best{0.756\,$\pm$\,0.034} & \best{0.753\,$\pm$\,0.035} & \best{0.755\,$\pm$\,0.067} & \second{0.803\,$\pm$\,0.053} & \best{0.739\,$\pm$\,0.035} & \second{0.746\,$\pm$\,0.015} & 0.736\,$\pm$\,0.040 & \best{0.802\,$\pm$\,0.040} \\
    \bottomrule
  \end{tabular*}
    \caption{Performance comparison on four seizure-detection datasets.
  Results are reported as mean $\pm$ standard deviation.
  Best results are in red and bold; second-best results are in blue; third-best results are underlined. V–L denotes vision–language.}
    \label{tab:main_results}
\end{table*}

\section{Experiments}

\subsection{Experimental Setup}

\paragraph{Datasets.}
We evaluate ViTexSZ on four EEG epilepsy datasets for seizure detection. Specifically, \textbf{Dual-SCS} is a clinical paired iEEG–scEEG dataset collected for subclinical seizure detection in collaboration with The \textbf{\textit{First Affiliated Hospital, Sun Yat-sen University.}} Informed consent was obtained from all 24 participating patients with epilepsy, and the data were fully de-identified before analysis. The scEEG recordings were acquired at 500 Hz using 32 channels, whereas the iEEG recordings were sampled at the same frequency with patient-specific channel configurations.
\textbf{Dual-CS} was collected at the same hospital but for clinical seizure detection. It contains paired iEEG--scEEG recordings from 39 patients. Both modalities were sampled at 256~Hz, with 32 scEEG channels and patient-specific iEEG channel configurations. The collection and research use of the Dual-SCS and Dual-CS datasets were approved by the Institutional Review Board, and all procedures were conducted in accordance with the Declaration of Helsinki~\cite{goodyear2007declaration}.
\textbf{CHB-MIT}~\cite{shoeb2010application} is a pediatric scEEG dataset from Boston Children's Hospital that contains long-term recordings from 24 cases. The recordings were acquired at 256~Hz using 23 bipolar channels and are accompanied by expert seizure annotations. 
\textbf{TUSZ}~\cite{obeid2016temple,shah2018temple} is a clinical scEEG dataset from Temple University Hospital. The recordings were acquired at 250~Hz using 19--21 standard channels following the international 10--20 system. We selected 238 patients with valid annotations, including 2,663 seizure events and corresponding background segments.

Overall, the collection and use of Dual-SCS and Dual-CS were approved by the Institutional Review Board of the First Affiliated Hospital of Sun Yat-sen University (Ethical Approval No. [2024]275). In addition, CHB-MIT and TUSZ were used in accordance with their original ethical approvals.

\paragraph{Data Preprocessing.}
The Dual-SCS and Dual-CS datasets were band-pass filtered between 1 and 80~Hz with an additional 50-Hz notch filter, whereas CHB-MIT and TUSZ retained their original configurations. After removing non-EEG channels and resampling where necessary, normal background segments were sampled from seizure-free periods within the same recordings. Their total duration was matched to that of the seizure intervals, with exclusion margins applied around seizure boundaries. Seizure and normal recordings were subsequently divided into 10-s ictal and non-ictal windows, respectively. Each window was rendered as a multichannel EEG image, with channels displayed in their original order using an average reference (AR) montage ~\cite{gu2026cerebragloss}, while the corresponding multichannel signals were directly used for the time-series models. The same window segmentation and channel configurations were maintained across both representations.

\paragraph{Evaluation Metrics.}
We report Accuracy (Acc), F1-score (F1), Sensitivity (Sens), and the area under the receiver operating characteristic curve (AUC). Sensitivity is particularly important for seizure screening because a false negative corresponds to a missed seizure event. Acc, F1, and Sens are computed at a validation-selected operating threshold, whereas AUC evaluates threshold-independent ranking ability. Therefore, these metrics may exhibit non-monotonic trade-offs. 

\paragraph{Baselines.}
We compare ViTexSZ against representative baselines spanning diverse time-series, visual, and vision-language categories.
\textbf{Time series-centric models.}
These methods directly process multivariate scEEG waveforms, including 
Transformer~\cite{vaswani2017attention}, temporal convolutional network (TCN)~\cite{bai2018tcn}, TimeMIL~\cite{chen2024timemil}, TeCh~\cite{yu2026tech}, ConvNeXt-Seizure~\cite{hogan2025scaling}, and MedGNN~\cite{fan2025medgnn}.
\textbf{Visual and vision-language models.}
These methods operate on the structured scEEG visual representations, including ResNet-18~\cite{he2016resnet}, ConvNeXt v2~\cite{woo2023convnextv2}, CLIP~\cite{radford2021clip}, BLIP~\cite{li2022blip}, SigLIP~\cite{zhai2023siglip}, and VIPEEGNet~\cite{sun2025automated}. For pretrained backbones, we fine-tune the corresponding encoder and task-specific classification head under the same data split and evaluation protocol.


\subsection{Experimental Results}

\definecolor{bestred}{RGB}{160,0,0}
\definecolor{secondblue}{RGB}{0,0,140}
\providecommand{\best}[1]{\textcolor{bestred}{\textbf{#1}}}
\providecommand{\second}[1]{\textcolor{secondblue}{#1}}
\providecommand{\third}[1]{\underline{#1}}


\paragraph{Performance Comparison.}
Table~\ref{tab:main_results} summarizes the performance on four seizure-detection datasets under a cross-subject setting using five patient-independent data splits. ViTexSZ achieves the highest accuracy on all four datasets and ranks first in 11 of the 16 dataset--metric combinations. The following observations can be concluded. (1) \textit{ViTexSZ performs particularly well on paired heterogeneous EEG data.} On Dual-CS, ViTexSZ achieves relative improvements of 12.9\% and 7.9\% over the corresponding second-best baselines in sensitivity and F1 score, respectively. These gains suggest that multi-channel alignment and vision--text knowledge transfer facilitate the recognition of seizure patterns across different channel configurations. (2) \textit{ViTexSZ is particularly effective for subclinical seizures without observable clinical manifestations.} Its accuracy on Dual-SCS exceeds its accuracies on Dual-CS, CHB-MIT, and TUSZ by 8.5, 9.5, and 11.2 percentage points, respectively. This result suggests that privileged iEEG supervision and token-level knowledge distillation help capture subtle electrographic patterns that may not manifest as observable clinical symptoms, thereby supporting SCS. (3) \textit{ViTexSZ generalizes well to large-scale scEEG datasets.} On TUSZ, which contains the largest cohort among the evaluated datasets, ViTexSZ ranks first in accuracy and AUC and second in F1 score among all compared methods. Together with its leading performance on CHB-MIT, these results demonstrate that ViTexSZ generalizes beyond the paired clinical datasets to large-scale scEEG settings. 


\begin{table}[t]
\centering
\footnotesize
\setlength{\tabcolsep}{4.2pt}
\renewcommand{\arraystretch}{1.0}
\newcommand{\abv}[1]{{\fontsize{7.8}{9}\selectfont #1}}
\newcommand{\abfirst}[1]{{\fontsize{7.8}{9}\selectfont\textbf{#1}}}
\newcommand{\absecond}[1]{{\fontsize{7.8}{9}\selectfont #1}}
\begin{tabular*}{\linewidth}{@{\extracolsep{\fill}}
  >{\centering\arraybackslash}m{42pt}
  >{\centering\arraybackslash}m{52pt}|
  *{4}{>{\centering\arraybackslash}m{23pt}}@{}}
\toprule
\textbf{Data} & \textbf{Variant} & \textbf{Acc} & \textbf{F1} & \textbf{Sens} & \textbf{AUC} \\
\midrule
\multirow{6}{*}{Dual-SCS}
 & w/o Text$_S$   & \absecond{\textcolor{DarkBlue}{0.8922}} & \absecond{\textcolor{DarkBlue}{0.8929}} & \abv{0.8929} & \absecond{\textcolor{DarkBlue}{0.9604}} \\
 & w/o Text$_T$   & \abv{0.8862} & \abv{0.8927} & \abv{0.9405} & \abv{0.9585} \\
 & w/o Teacher    & \abv{0.8443} & \abv{0.8539} & \abv{0.9048} & \abv{0.8876} \\
 & w/o iEEG       & \abv{0.8683} & \abv{0.8791} & \absecond{\textcolor{DarkBlue}{0.9524}} & \abv{0.9005} \\
 & w/o MCA        & \abv{0.8683} & \abv{0.8642} & \abv{0.8333} & \abv{0.9562} \\
\cmidrule(l){2-6}
 & \textbf{ViTexSZ} & \abfirst{\textcolor{DarkRed}{0.8982}} & \abfirst{\textcolor{DarkRed}{0.9050}} & \abfirst{\textcolor{DarkRed}{0.9643}} & \abfirst{\textcolor{DarkRed}{0.9667}} \\
\midrule
\multirow{6}{*}{Dual-CS}
 & w/o Text$_S$   & \abv{0.8193} & \abv{0.8148} & \abv{0.7765} & \abv{0.8704} \\
 & w/o Text$_T$   & \absecond{\textcolor{DarkBlue}{0.8133}} & \absecond{\textcolor{DarkBlue}{0.8187}} & \absecond{\textcolor{DarkBlue}{0.8235}} & \absecond{\textcolor{DarkBlue}{0.8786}} \\
 & w/o Teacher    & \abv{0.7771} & \abv{0.7811} & \abv{0.7765} & \abv{0.8365} \\
 & w/o iEEG       & \abv{0.7952} & \abv{0.7976} & \abv{0.7882} & \abv{0.8559} \\
 & w/o MCA        & \abv{0.8012} & \abv{0.8114} & \abfirst{\textcolor{DarkRed}{0.8353}} & \abv{0.8680} \\
\cmidrule(l){2-6}
 & \textbf{ViTexSZ} & \abfirst{\textcolor{DarkRed}{0.8193}} & \abfirst{\textcolor{DarkRed}{0.8214}} & \abv{0.8118} & \abfirst{\textcolor{DarkRed}{0.8821}} \\
\bottomrule
\end{tabular*}
\caption{Ablation study on the first fold of each dataset. Removing the teacher results in the largest performance drop.
}
\label{tab:ablation}
\end{table}
\vspace{-5pt}

\paragraph{Ablation Study.}
Table~\ref{tab:ablation} evaluates the contribution of each component on Dual-SCS and Dual-CS. (1) \textit{The heterogeneous teacher provides the primary supervision for the lightweight student.} 
Removing the teacher decreases AUC by 7.9 percentage points.
This degradation indicates that the student benefits from both multimodal teacher representations and token-level privileged knowledge transfer. 
(2) \textit{Multi-channel alignment is particularly important for seizure sensitivity.} 
Without MCA, sensitivity decreases by 13.1\% in the former, while accuracy, F1, and AUC decline in the latter. This suggests that mapping heterogeneous channel representations into a unified token space helps preserve seizure-related evidence across datasets.
(3) \textit{Intracranial EEG contributes complementary seizure evidence.} 
Removing iEEG reduces AUC by 6.6\% and 2.6\%.
This result suggests that higher-fidelity intracranial recordings improve the separability of seizure patterns during teacher training. (4) \textit{Clinical prompts provide complementary semantic guidance in both branches}, which indicate that clinical semantics primarily strengthen the teacher representation and further condition the student during seizure detection.

\begin{figure}[t]
    \centering
    \hspace*{-2pt}%
    \includegraphics[width=\linewidth,trim=4 35 0 0,clip]{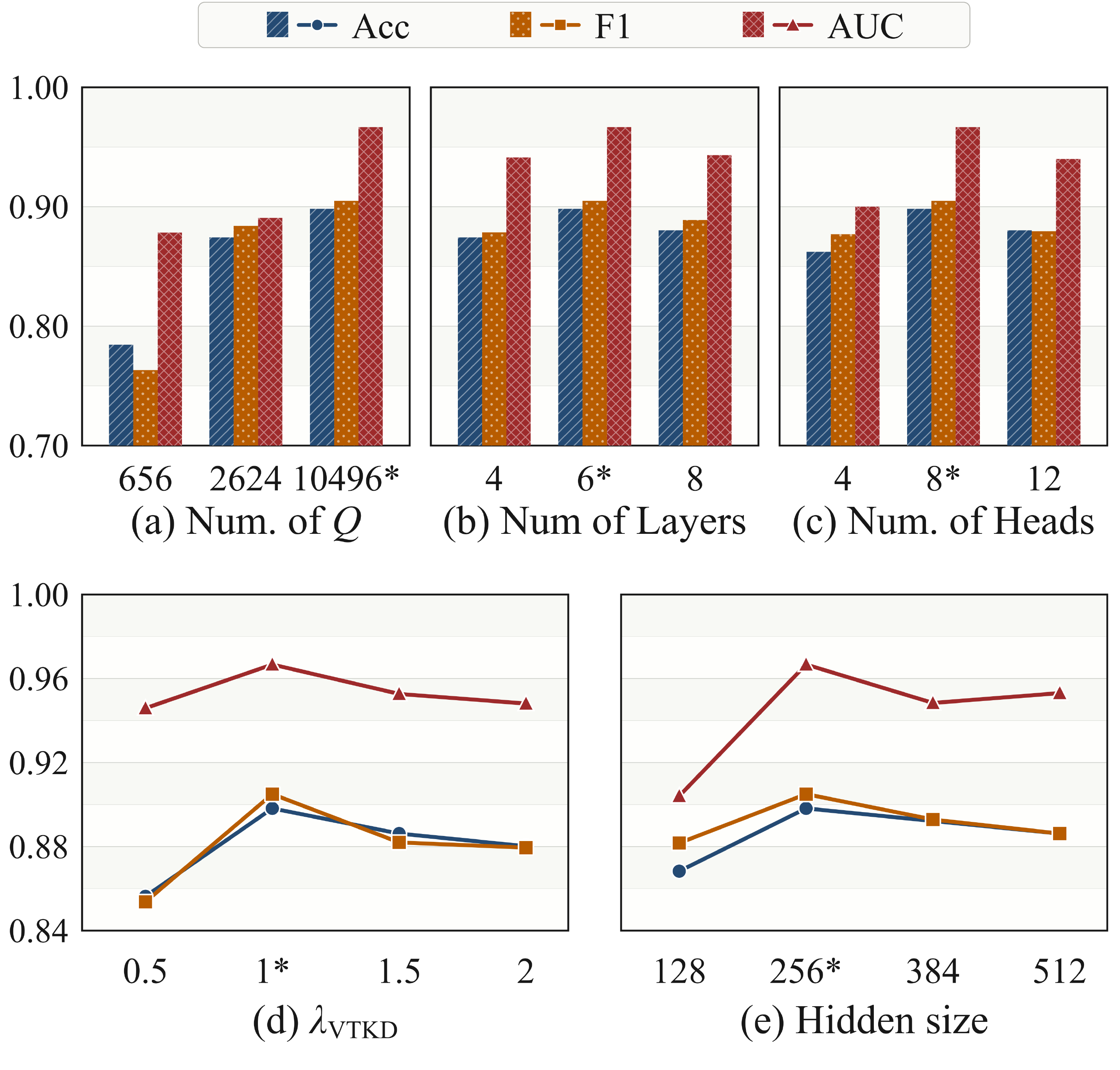}
    \vspace{-0.5cm}
    \caption{Hyperparameter sensitivity to (a) query count, (b) Transformer depth, (c) attention heads, (d) distillation weight, and (e) hidden dimension.}
    \vspace{-0.5cm}
    \label{fig:Sensitivity_analysis}
\end{figure}

\paragraph{Hyperparameter Analysis.}

As shown in Figure~\ref{fig:Sensitivity_analysis}, we vary each hyperpparameter independently while keeping the others fixed on Dual-SCS. Increasing the number of queries $N_q$ from 656 to 10,496 improves Acc, F1, and AUC, suggesting that a larger query set benefits cross-modal aggregation. The best results are obtained with 6 Transformer layers and 8 attention heads, while further increasing either provides no clear gain. Performance is also maximized at $\lambda_{\mathrm{VTKD}}=1.0$, indicating a suitable balance between supervised learning and privileged knowledge transfer, and at a hidden dimension of $D=256$. We use these settings as the default configuration in all experiments. 

\paragraph{T-SNE Visualization.}

\begin{figure}[t]
    \centering
    \hspace*{-2pt}%
    \includegraphics[width=\linewidth,trim=0 6 0 0,clip]{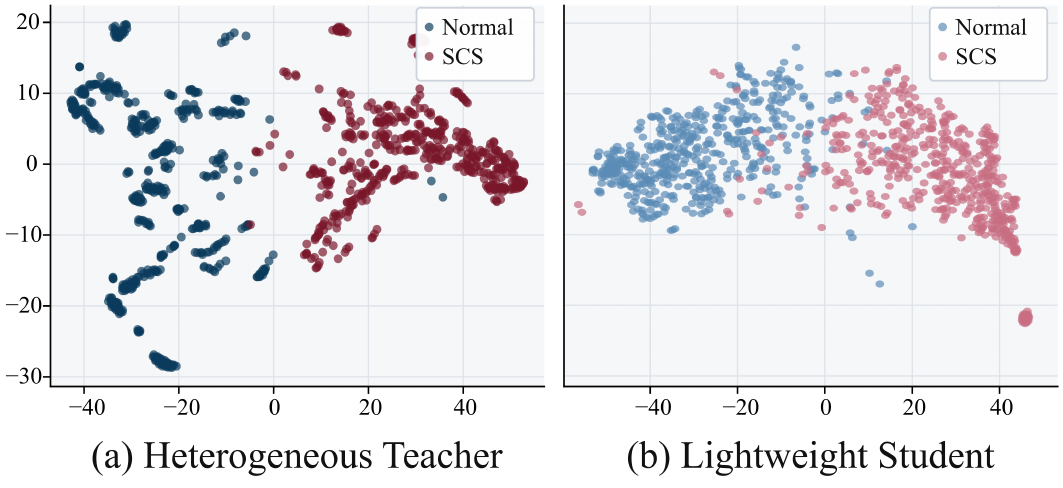}
    \vspace{-0.5cm}
    \caption{T-SNE visualization of final representations.}
    \label{fig:tsne}
\end{figure}

Figure~\ref{fig:tsne} visualizes the feature distributions learned by the heterogeneous teacher and the lighweight student for subclinical seizure detection. By combining pretrained multimodal embeddings with cross-modal fusion of scEEG and iEEG, the heterogeneous teacher forms well-separated clusters for SCS and normal samples. After the vision-text knowledge distillation, the lighweight student retains a similarly discriminative feature structure using only scEEG during real detection. This result shows that seizure-relevant multimodal knowledge is effectively transferred to the lightweight student. 


\paragraph{Case Study.}
We evaluate ViTexSZ through continuous monitoring of a previously unseen, expert-annotated 3-hour scEEG recording provided by a tertiary general hospital for real-time subclinical seizure detection. The recording is sequentially divided into non-overlapping 10-s windows, and ViTexSZ generates a seizure prediction for each window in chronological order. ViTexSZ achieves an AUPRC of 0.83 and an F1 score of 0.85, demonstrating robust detection performance in this realistic online setting. Online inference requires approximately 9.8~ms per window, further demonstrating the feasibility of ViTexSZ for real-time continuous monitoring.

\begin{figure}[t]
    \centering
    \includegraphics[width=1\linewidth]{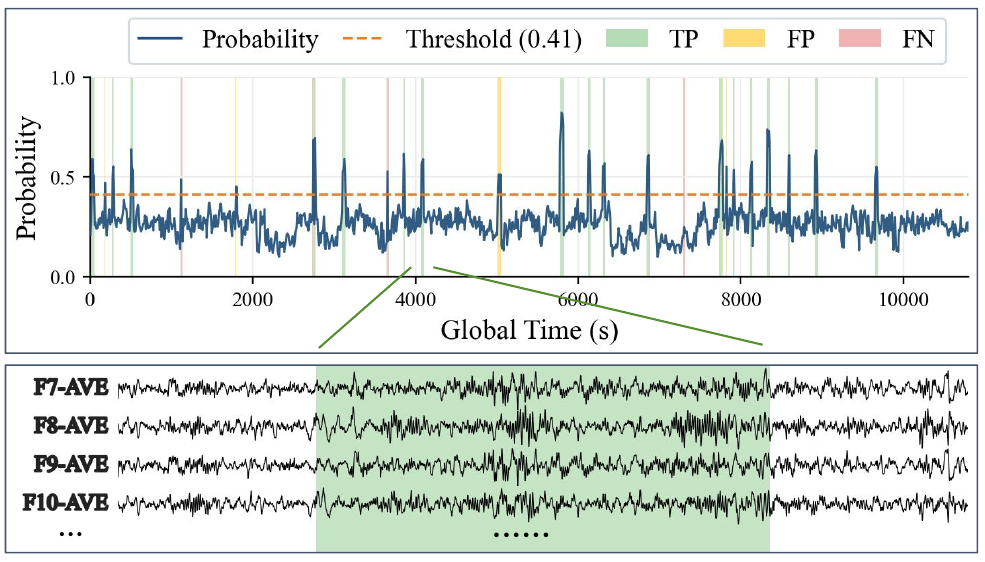}
    \vspace{-0.5cm}
    \caption{Real-time seizure detection on clinical scEEG.}
    \vspace{-0.5cm}
    \label{fig:Case_study}
\end{figure}

\section{Conclusion}

In this study, we introduce \textbf{ViTexSZ}, a heterogeneous vision--text knowledge distillation framework for EEG seizure detection. ViTexSZ converts heterogeneous EEG recordings into waveform images and employs multi-channel alignment to project source-dependent representations into a unified token space. A heterogeneous teacher associates seizure-related visual patterns with clinical semantics through a multimodal large language model, while token-level knowledge distillation transfers these vision--text representations to a lightweight student. Experiments on heterogeneous EEG datasets demonstrate the effectiveness and generalizability of ViTexSZ, particularly for subclinical seizure detection. Future work will explore a seizure foundation model with a shared prediction head for cross-center detection.

\bibliography{aaai2027}


\end{document}